\documentclass[letterpaper]{article} 

\usepackage{main}
\usepackage{times}
\usepackage{helvet}
\usepackage{courier}
\usepackage[hyphens]{url}
\usepackage{graphicx}
\usepackage{natbib}
\usepackage{caption}

\usepackage{amsmath}
\usepackage{amssymb}
\usepackage{multirow}
\usepackage{booktabs}
\usepackage{xspace}
\usepackage[table]{xcolor}

\usepackage{algorithm}
\usepackage{algorithmic}

\usepackage{newfloat}
\usepackage{listings}
\DeclareCaptionStyle{ruled}{labelfont=normalfont,labelsep=colon,strut=off}
\floatstyle{ruled}
\newfloat{listing}{tb}{lst}{}
\floatname{listing}{Listing}

\usepackage{xcolor}

\definecolor{link}{RGB}{255,128,134}
\definecolor{la}{RGB}{0,158,57}
\usepackage[pagebackref=false,breaklinks,colorlinks,allcolors=link]{hyperref}

\usepackage{cleveref}

\title{
\vspace{-0.7cm}
    \includegraphics[width=0.05\linewidth]{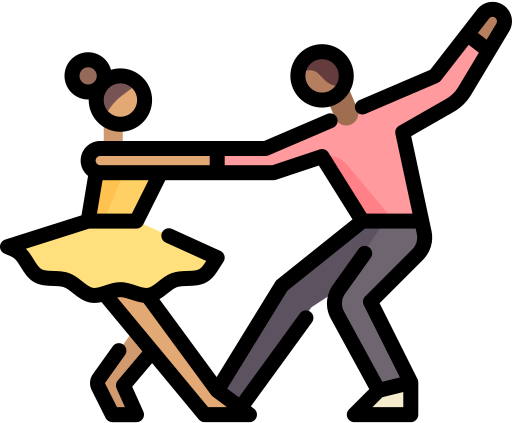}~La La LiDAR: \underline{La}rge-Scale \underline{La}yout Generation from \underline{LiDAR} Data
}

\author{
Youquan Liu\textsuperscript{\rm 1}\quad 
Lingdong Kong\textsuperscript{\rm 2}\quad
Weidong Yang\textsuperscript{\rm 1,}\thanks{Corresponding authors.}\quad
Xin Li\textsuperscript{\rm 3}\quad
Ao Liang\textsuperscript{\rm 2}\\ 
Runnan Chen\textsuperscript{\rm 4,}\footnotemark[1]\quad
Ben Fei\textsuperscript{\rm 5,}\footnotemark[1]\quad
Tongliang Liu\textsuperscript{\rm 4,}\footnotemark[1]
}
\affiliations{
\textsuperscript{\rm 1}Fudan University \quad
\textsuperscript{\rm 2}National University of Singapore \quad
\textsuperscript{\rm 3}Shanghai AI Laboratory \\
\textsuperscript{\rm 4}University of Sydney \quad
\textsuperscript{\rm 5}The Chinese University of Hong Kong
}

\begin{document}

\maketitle

\begin{abstract}
Controllable generation of realistic LiDAR scenes is crucial for applications such as autonomous driving and robotics. While recent diffusion-based models achieve high-fidelity LiDAR generation, they lack explicit control over foreground objects and spatial relationships, limiting their usefulness for scenario simulation and safety validation. To address these limitations, we propose \underline{\textbf{La}}rge-scale \underline{\textbf{La}}yout-guided \underline{\textbf{LiDAR}} generation model (``\textit{\textbf{La La LiDAR}}''),  a novel layout-guided generative framework that introduces semantic-enhanced scene graph diffusion with relation-aware contextual conditioning for structured LiDAR layout generation, followed by foreground-aware control injection for complete scene generation. This enables customizable control over object placement while ensuring spatial and semantic consistency. To support our structured LiDAR generation, we introduce Waymo-SG and nuScenes-SG, two large-scale LiDAR scene graph datasets, along with new evaluation metrics for layout synthesis. Extensive experiments demonstrate that La La LiDAR achieves state-of-the-art performance in both LiDAR generation and downstream perception tasks, establishing a new benchmark for controllable 3D scene generation.
\end{abstract}
\begin{figure}[!ht]
    \centering
    \includegraphics[width=0.95\linewidth]{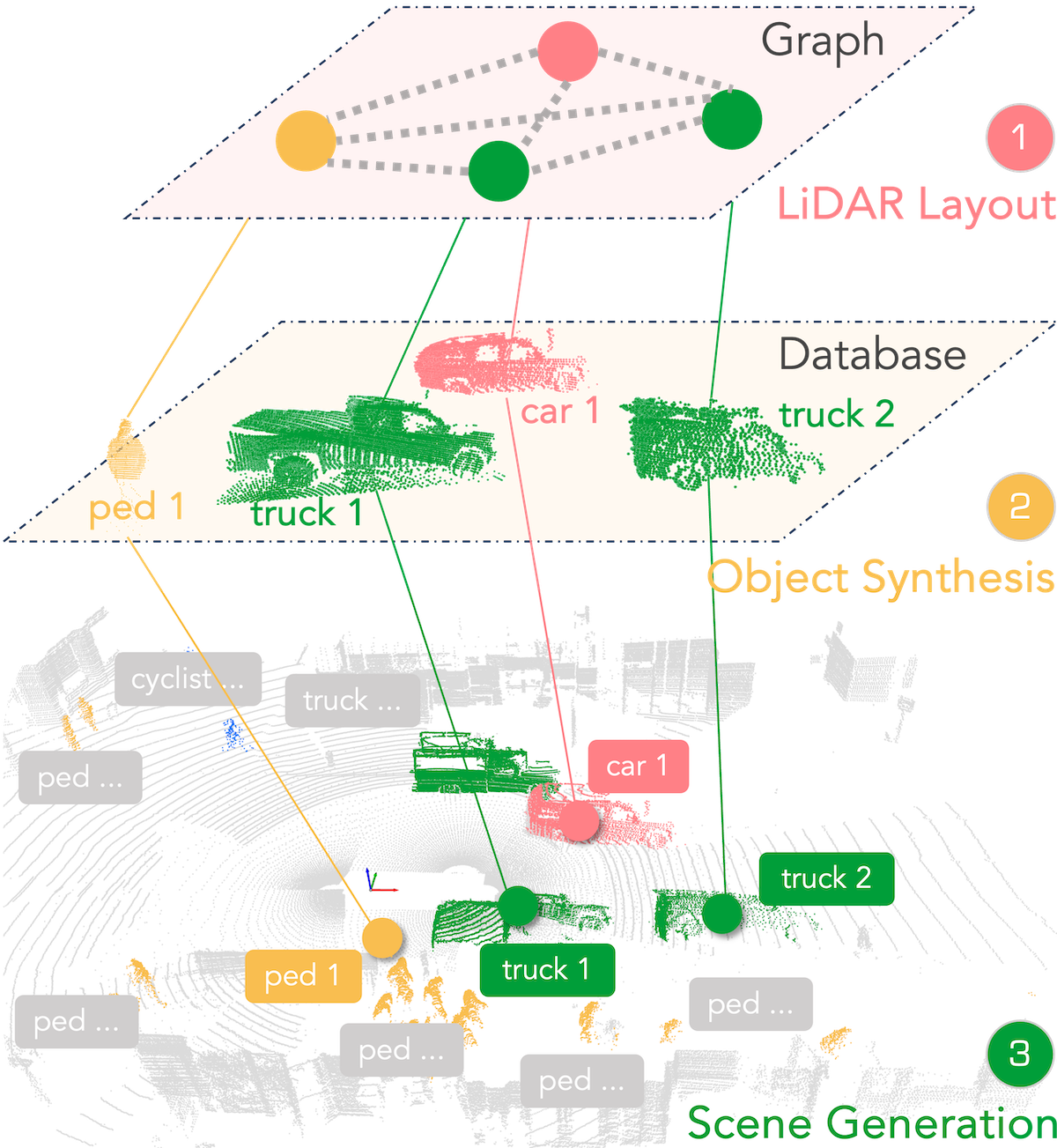}
    \vspace{-0.15cm}
    \caption{
    Motivation of \textbf{customizable LiDAR scene generation} from ``\textit{\textbf{La La LiDAR}}''. Our framework consists of \textbf{three key stages}: \textbf{1)} \textbf{LiDAR layout generation} using scene graphs, where nodes represent objects and edges capture their spatial relationships; \textbf{2)} \textbf{foreground point cloud synthesis}, either by retrieving from a database or by generating conditioned on layout parameters; and \textbf{3)} \textbf{foreground-conditioned scene generation}, where synthesized foreground serves as conditioning to generate the complete scene with realistic environmental context. This hierarchical approach enables fine-grained control over foreground object placement while maintaining overall scene coherence.
    }
    \label{fig:teaser}
    \vspace{-0.3cm}
\end{figure}

\section{Introduction}
\label{sec:intro}

Generating 3D scenes has emerged as a critical technology for a wide spectrum of real-world applications, including autonomous driving, AR/VR, and robotics  \cite{muhammad2022survey, bian2025dynamiccity,lee2024semcity}. In particular, for autonomous driving systems, the quality, diversity, and controllability of 3D training data directly impact perception capabilities and generalization performance \cite{wilson2022carlasc,zheng2024lidar4d,xie2025citydreamer4d}. Traditional data collection methods, however, face substantial limitations in terms of cost, scalability, and coverage of rare scenarios, making synthetic data generation an increasingly attractive alternative. Recent advancements in generative models, especially diffusion-based approaches \cite{ddpm,latentdiffusion}, have demonstrated remarkable success in high-fidelity image synthesis. This progress has naturally extended to 3D representations, yielding promising results for LiDAR point cloud generation in driving scenarios \cite{hu2024rangeldm,nakashima2021learning,nunes2024scaling,ran2024lidm,zyrianov2022lidargen}.

Despite notable advances in unconditional LiDAR scene generation, current approaches suffer from a critical limitation: the lack of customizable control over foreground object composition and arrangement. This deficiency is particularly significant in autonomous driving scenarios, where foreground objects such as vehicles, pedestrians, and other traffic participants constitute only a small fraction (approximately $6.29\%$, measured on nuScenes) of the total point cloud. Although these foreground elements are crucial for downstream perception tasks, existing methods like LiDARGen \cite{zyrianov2022lidargen} process foreground and background elements uniformly, offering minimal control over object placement and inter-object spatial relationships. This uniform treatment limits the applicability of these generative models for driving simulation and long-tail scenario synthesis.

To address these limitations, we propose \textbf{La La LiDAR} (Large-scale Layout-guided LiDAR) as shown in \Cref{fig:teaser}, a novel framework for fine-grained controllable LiDAR scene generation that explicitly models spatial relationships between foreground objects through structured scene graph representations. Our approach comprises three main steps: \textbf{1)} Structured layout generation: synthesizing LiDAR layouts via a scene graph-based diffusion model to ensure spatial and semantic coherence. \textbf{2)} Foreground point cloud synthesis: retrieving or generating object-level point clouds conditioned on the layout. \textbf{3)} Foreground-conditioned scene generation: leveraging the synthesized foreground as control signals to generate a complete LiDAR scene with realistic environmental context. By integrating graph-based spatial reasoning and diffusion-based generative modeling, our method enables fine-grained customization of scene composition while preserving the realism of large-scale LiDAR environments. The contributions of our work include:
\begin{itemize}
\item {
    We introduce \textbf{La La LiDAR}, a comprehensive LiDAR layout generation framework that leverages scene graphs to capture semantic and spatial relationships among foreground objects. This enables customizable scene composition while maintaining physical plausibility.
}
\item {
    We propose a novel Foreground-aware Control Injector (FCI) that effectively bridges layout information with the scene generation process, ensuring accurate representation and spatial coherence of foreground elements within the broader environmental context.
}
\item {
    We construct two large-scale LiDAR scene graph datasets, \textbf{Waymo-SG} and \textbf{nuScenes-SG}, along with specialized evaluation metrics tailored for LiDAR layout generation, establishing new benchmarks for future research in LiDAR layout synthesis.
}
\item {
    Our analyses prove superior performance of our approach in LiDAR layout generation, LiDAR scene generation, and multiple downstream perception tasks, including LiDAR semantic segmentation, 3D object detection, and scene completion.}
\end{itemize}

\noindent Extensive experiments under diverse setups demonstrate that La La LiDAR not only produces high-fidelity scenes but also provides unprecedented control over scene composition. By explicitly modeling foreground objects and their relationships, our method addresses a critical need in the development and evaluation of autonomous driving systems, enabling the generation of more diverse, realistic, and controllable driving scenarios.
\section{Related Work}
\label{sec:related_work}
\noindent\textbf{LiDAR Scene Generation.} LiDARGen~\cite{zyrianov2022lidargen} pioneered diffusion-based LiDAR generation using range and reflectance data. UltraLiDAR \cite{xiong2023ultralidar} integrates voxelized LiDAR point cloud with a VQ-VAE framework \cite{van2017neural}, thereby enabling efficient LiDAR data generation and completion. R2DM~\cite{nakashima2024r2dm} improved generation quality with advanced denoising networks, while LiDM~\cite{ran2024lidm} focused on preserving geometric structures. RangeLDM~\cite{hu2024rangeldm} emphasized real-time generation, and Text2LiDAR~\cite{wu2024text2lidar} introduced text-guided synthesis for semantic control. However, existing methods treat foreground and background uniformly, limiting control over object placement and spatial relationships. To address this, we propose a layout-guided LiDAR generation framework for more controllable and diverse LiDAR scene generation.

\noindent\textbf{3D Generation from Scene Layouts.}
While extensively studied in indoor environments, scene layout-based generation remains underexplored for outdoor settings. Methods like Graph-to-3D~\cite{dhamo2021graph-to-3d} and CommonScenes~\cite{zhai2023commonscenes} utilize scene graphs for spatially coherent room generation, while EchoScene~\cite{zhai2024echoscene} further improves layout consistency through information echo mechanisms. We introduce the first scene graph-based approach for outdoor LiDAR generation, explicitly modeling foreground objects and their spatial relationships while proposing tailored evaluation metrics for autonomous driving scenarios.

\noindent\textbf{Downstream Applications.}
LiDAR-based perception plays a crucial role in autonomous driving applications~\cite{li2022homogeneous,kong2023robo3D,liu2024m3net,liu2023uniseg,li2023logonet}. For LiDAR semantic segmentation, methods like SPVCNN~\cite{tang2020searching} and MinkUNet~\cite{choy2019minkowski} leverage sparse convolutions, while semi-supervised approaches such as MeanTeacher \cite{tarvainen2017meanteacher} and LaserMix~\cite{kong2023laserMix, kong2025lasermix++} improve data efficiency. 3D object detection models like CenterPoint~\cite{yin2021centerpoint} and SECOND~\cite{second} localize traffic participants using bird’s-eye view or voxelized representations, directly impacting navigation safety. LiDAR completion tackles occlusion, with diffusion models \cite{zhao2025diffusion,zhang2024distilling, martyniuk2025lidpm} like Repaint \cite{lugmayr2022repaint} achieving superior reconstruction via learned priors. Our work enhances these applications by generating high-quality LiDAR data with controllable foreground elements, improving robustness in segmentation, detection, and completion.
\section{Methodology}
\label{sec:method}
We propose a layout-guided LiDAR generation framework comprising two stages: (i) layout generation and foreground synthesis, (ii) foreground-aware scene generation. This design enables customized control over foreground objects while ensuring spatial coherence, addressing the limitations of existing LiDAR scene generation approaches.

\begin{figure*}[t]
    \centering
    \includegraphics[width=\textwidth]{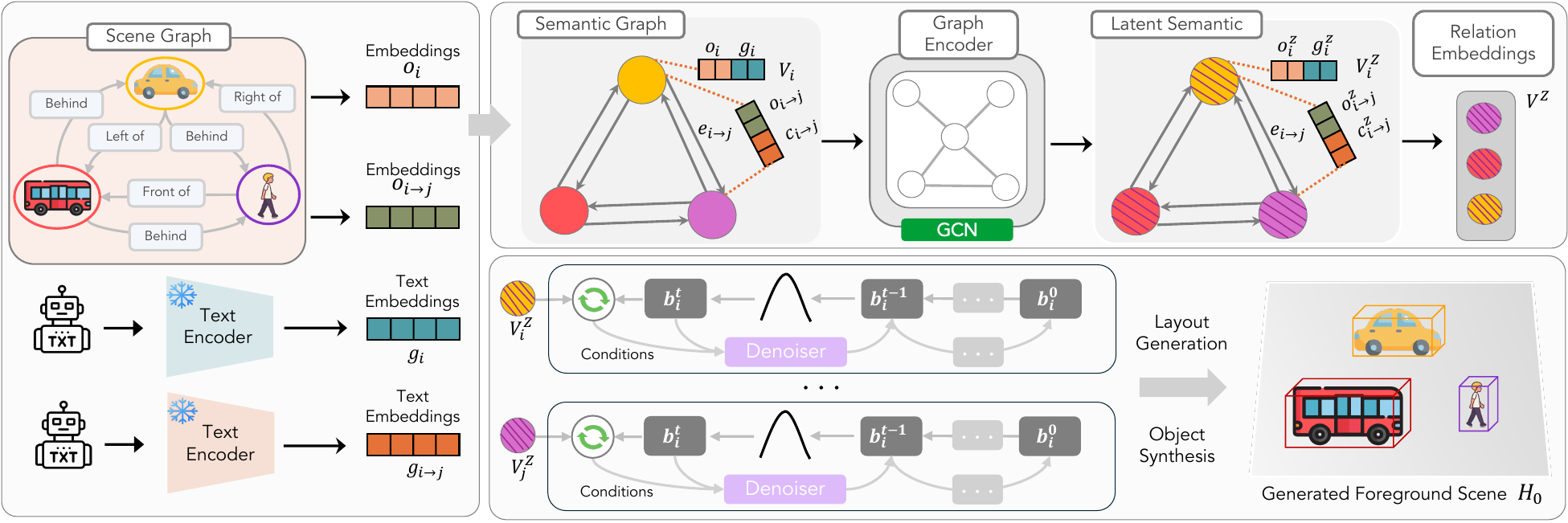}
    \vspace{-0.5cm}
    \caption{The proposed \textbf{LiDAR point cloud layout generation} framework. Our approach begins with scene graph construction, establishing both node embeddings ($o_i$) and edge embeddings ($o_{i\rightarrow j}$) to capture spatial relationships. These are enhanced with semantic features from a CLIP text encoder ($g_i$, $g_{i\rightarrow j}$), creating a comprehensive semantic graph. Graph Encoder then processes this information to produce a latent semantic graph with enriched node representations ($V_i^Z$). During the diffusion process, layout states ($b_i^t$) are iteratively refined through a denoising network that incorporates time-dependent contextual conditioning ($\mathcal{C}_t$), which dynamically aggregates graph features at each timestep. This ensures consistent spatial relationships throughout the denoising process. The final stage synthesizes and places appropriate foreground points according to the generated layout.}
    \label{fig:layout_framework}
\end{figure*}

\subsection{Preliminaries}
\textbf{LiDAR Representation.}  
A LiDAR point cloud is defined as $\mathcal{P} = \{ (\mathbf{p}^i, \mathbf{e}^i) \mid i = 1, ..., N \}$, where each point $\mathbf{p}^i \in \mathbb{R}^3$ represents the 3D coordinates $(p_x^i, p_y^i, p_z^i)$ and $\mathbf{e}^i \in \mathbb{R}^L$ denotes auxiliary attributes such as intensity. Following prior works~\cite{zyrianov2022lidargen,wu2024text2lidar}, we adopt a spherical projection \cite{milioto2019rangenet} to convert $\mathcal{P}$ into a structured range image $X \in \mathbb{R}^{H \times W \times 2}$ for efficient processing. The projection process $\Pi: \mathbb{R}^3 \mapsto \mathbb{R}^2$ is defined as follows:
\begin{equation}
\left( \begin{array}{c} u \\ v \end{array} \right) = 
\left( \begin{array}{c} 
\frac{1}{2} \left[ 1 - \frac{\arctan(p_y^i, p_x^i)}{\pi} \right] W \\
\left[ 1 - \frac{\arcsin(p_z^i / d) + f_{\text{up}}}{f} \right] H
\end{array} \right)~,
\end{equation}
where $d = \|\mathbf{p}^i\|_2$ is the depth, $f = f_{\text{up}} + f_{\text{down}}$ is the vertical field-of-view, and $(H, W)$ denote the vertical and horizontal resolutions. Each pixel in $X$ stores the depth and intensity.

\noindent\textbf{Conditional Diffusion Models.}  
We adopt the denoising diffusion probabilistic model (DDPM)~\cite{ddpm} conditioned on structured inputs. The model learns to predict the Gaussian noise $\epsilon$ added to clean data $\mathbf{x}_0$ through a noise prediction network $\epsilon_\theta$. The training objective minimizes:
\begin{equation}
\mathcal{L} = \mathbb{E}_{t, \mathbf{x}_0, \epsilon \sim \mathcal{N}(0, \mathbf{I})} \left[ \| \epsilon - \epsilon_\theta(\mathbf{x}_t, t, c) \|_2^2 \right]~,
\end{equation}
where $\mathbf{x}_t$ is the noisy sample at timestep $t$, and $c$ denotes a conditioning signal such as semantic layout or foreground. This framework enables controllable generation, which we exploit in both layout and scene synthesis stages.

\subsection{LiDAR Layout Generation}
Achieving realistic and controllable LiDAR scene generation requires explicitly modeling the spatial arrangement of foreground objects. To this end, we introduce a structured layout generation framework in \Cref{fig:layout_framework}, which leverages scene graphs to represent semantic and spatial relations among objects and guides the synthesis of object layouts.

\begin{figure}[t]
    \centering
    \includegraphics[width=\linewidth]{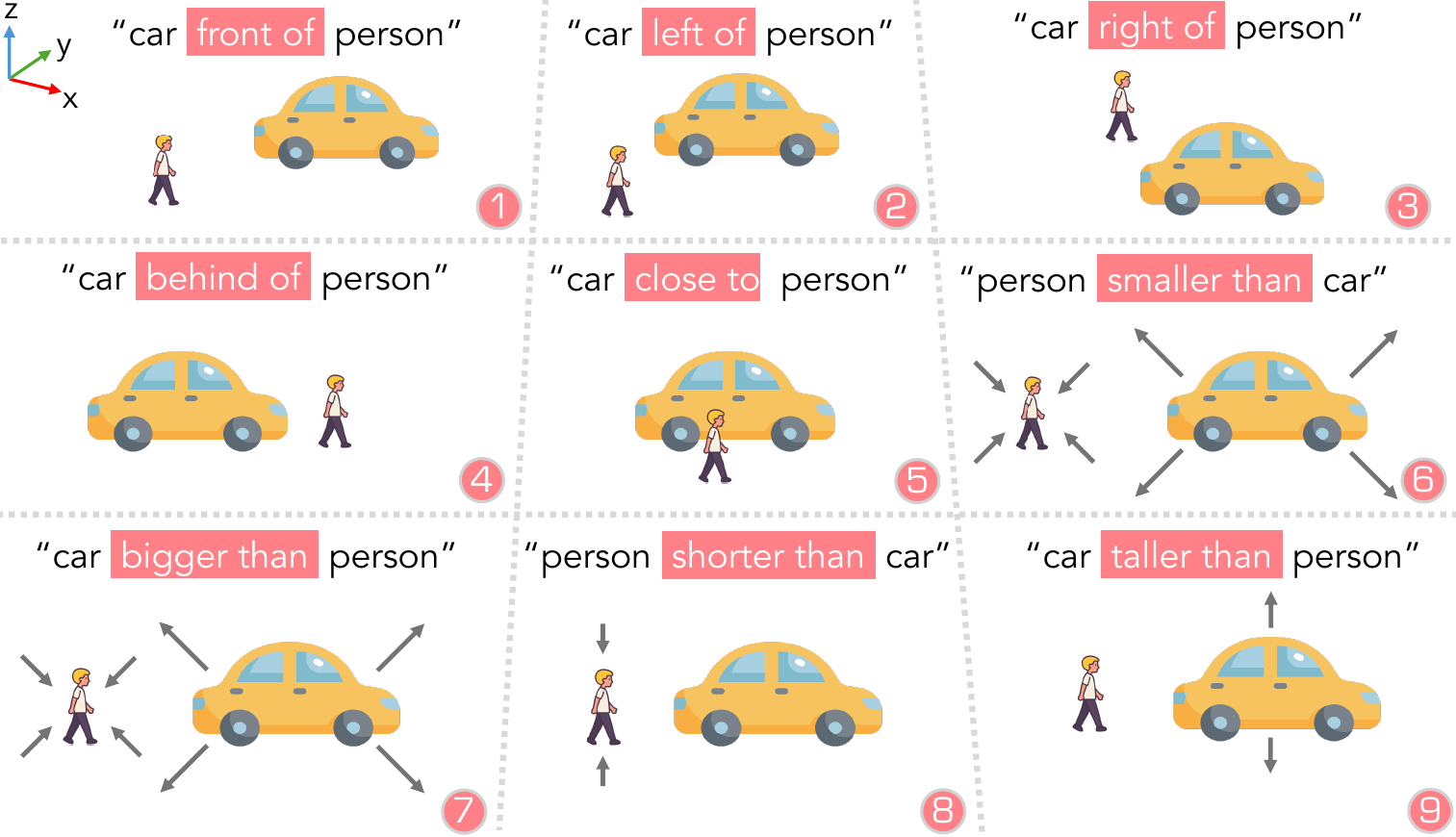}
    \vspace{-0.55cm}  
    \caption{The schematic definition of the nine \textbf{relationships} (foreground objects) in our LiDAR scene graph formulation.}
    \label{fig:scene_graph}
\end{figure}

\noindent\textbf{LiDAR Scene Graph Construction.}
We represent each LiDAR scene as a directed graph $\mathcal{G} = (\mathcal{V}, \mathcal{E})$, where nodes $v_i \in \mathcal{V}$ denote foreground objects, and edges $e_{i \to j} \in \mathcal{E}$ capture pairwise spatial or semantic relations.
We embed $v_i$ and $e_{i \to j}$ in the scene graph to derive node embeddings \( o_i \) and edge embeddings \( o_{i \to j} \). Each node and edge is labeled with class attributes, i.e., $c_i^{\mathrm{node}} \in \mathcal{C}^{\mathrm{node}}$ and $c_{i \to j}^{\mathrm{edge}} \in \mathcal{C}^{\mathrm{edge}}$, respectively. Each object layout $b_i$ is parameterized by a 3D bounding box in the ego-vehicle frame, including $(x, y, z)$ position, $(l, h, w)$ size, and yaw $\theta$. All parameters are normalized, and $\theta$ is encoded using sine-cosine to preserve angular continuity.
Due to the lack of existing LiDAR scene graph benchmarks, we construct two datasets: \textbf{Waymo-SG} and \textbf{nuScenes-SG}, derived from Waymo Open~\cite{sun2020waymo} and nuScenes~\cite{fong2022panoptic-nuscenes} datasets, respectively. These graphs encode nine relation types (see Figure~\ref{fig:scene_graph}), including spatial (\textit{e.g.}, \textit{front of}, \textit{left of}) and comparative (\textit{e.g.}, \textit{bigger than}, \textit{taller than}) relations. See the Appendix for additional dataset details.

\noindent\textbf{Graph-Based Semantic Encoding.}
To model structured spatial priors, we represent the input scene as a semantic graph and process it with a triplet Graph Convolutional Network (GCN). Following~\cite{johnson2018image,zhai2023commonscenes}, our GCN iteratively updates node and edge features by aggregating contextual messages. At each layer $k$, node $v_i$ and edge $e_{i \to j}$ are updated as:
\begin{equation}
\begin{aligned}
    &(\alpha_{v_{i}}^{(k)}, \beta_{e_{i\to j}}^{(k+1)}, \alpha_{v_{j}}^{(k)}) = \text{MLP}_1(\beta_{v_{i}}^{(k)}, \beta_{e_{i\to j}}^{(k)}, \beta_{v_{j}}^{(k)})~, \\
    &\beta_{v_{i}}^{(k+1)} = \alpha_{v_{i}}^{(k)} + \text{MLP}_2\left(\texttt{Avg}\left(\alpha_{v_j}^{(k)} \mid v_j \in N_{\mathcal{G}}(v_i)\right)\right)~,
\end{aligned}
\label{eq:gcn}
\end{equation}
where $\text{MLP}_1$, $\text{MLP}_2$ are multi-layer perceptrons, and $N_{\mathcal{G}}(v_i)$ denotes neighbors of $v_i$ in the scene graph.

To enhance semantic alignment, we leverage CLIP~\cite{radford2021clip} encoder $E_{\mathrm{clip}}$ to encode nodes and edges in a unified language-vision space. Given an object class $c_i^{\mathrm{node}}$, we compute $g_i = E_{\mathrm{clip}}(c_i^{\mathrm{node}})$. Relations are embedded via templated prompts (\textit{e.g.}, ``Car [left of] Pedestrian''), yielding $g_{i \to j} = E_{\mathrm{clip}}(c_i^{\mathrm{node}}, c_{i \to j}^{\mathrm{edge}}, c_j^{\mathrm{node}})$. These features are concatenated with learnable embeddings to form enriched node descriptors: $V_i = \mathrm{Concat}(g_i, o_i)$. The initial GCN inputs are set as $(\beta_{v_i}^{(0)}, \beta_{e_{i\to j}}^{(0)}, \beta_{v_j}^{(0)}) = (V_i, e_{i \to j}, V_j)$. After $K$ GCN layers, we obtain latent node embeddings $\{V_i^z\}$ that encode both semantic class and relational structure. These embeddings serve as the foundation for the subsequent layout diffusion process.

\noindent\textbf{Layout Diffusion.}
We formulate layout generation as a conditional diffusion process, where each foreground object node \( v_i \) is associated with a time-dependent noisy layout state \( b_i^t \in \mathbb{R}^8 \). To enable relationally grounded denoising, we construct a node-wise input embedding by concatenating its semantic feature \( V_i^z \), the temporal encoding \( \pi(t) \), and the noisy layout \( b_i^t \):
\begin{equation}
    \tilde{V}_i^t = \mathrm{Concat}(V_i^z, \pi(t), b_i^t)~.
\end{equation}

\noindent These features are processed by a GCN over the scene graph \( \mathcal{G} = (\mathcal{V}, \mathcal{E}) \), producing a time-varying contextual embedding \( \mathcal{C}_t \) that aggregates relational priors across the graph. Given the diffusion schedule coefficients $\alpha_t = 1 - \beta_t$ and $\bar{\alpha}_t = \prod_{s=1}^{t} \alpha_s$, the reverse step updates each node's layout as:
\begin{equation}
    b_i^{t-1} = \frac{1}{\sqrt{\alpha_t}} \left( b_i^t - \frac{1-\alpha_t}{\sqrt{1-\bar{\alpha}_t}} \epsilon_\theta(b_i^t, t, \mathcal{C}_t) \right) + \sigma_t \mathbf{z}~,
\end{equation}
where $\mathbf{z} \sim \mathcal{N}(0, I)$ for $t > 1$, and $\mathbf{z} = 0$ otherwise. The denoising network \( \epsilon_\theta \) is implemented as a cross-attention transformer, shared across nodes and conditioned on \( \mathcal{C}_t \). Starting from Gaussian noise \( b_i^T \sim \mathcal{N}(0, I) \), the model progressively refines object layouts over $T$ steps. By incorporating dynamically aggregated graph context, our model avoids node-level isolation~\cite{zhai2024echoscene} and enforces globally consistent spatial configurations during generation.

\begin{figure*}[t]
    \centering
    \includegraphics[width=\textwidth]{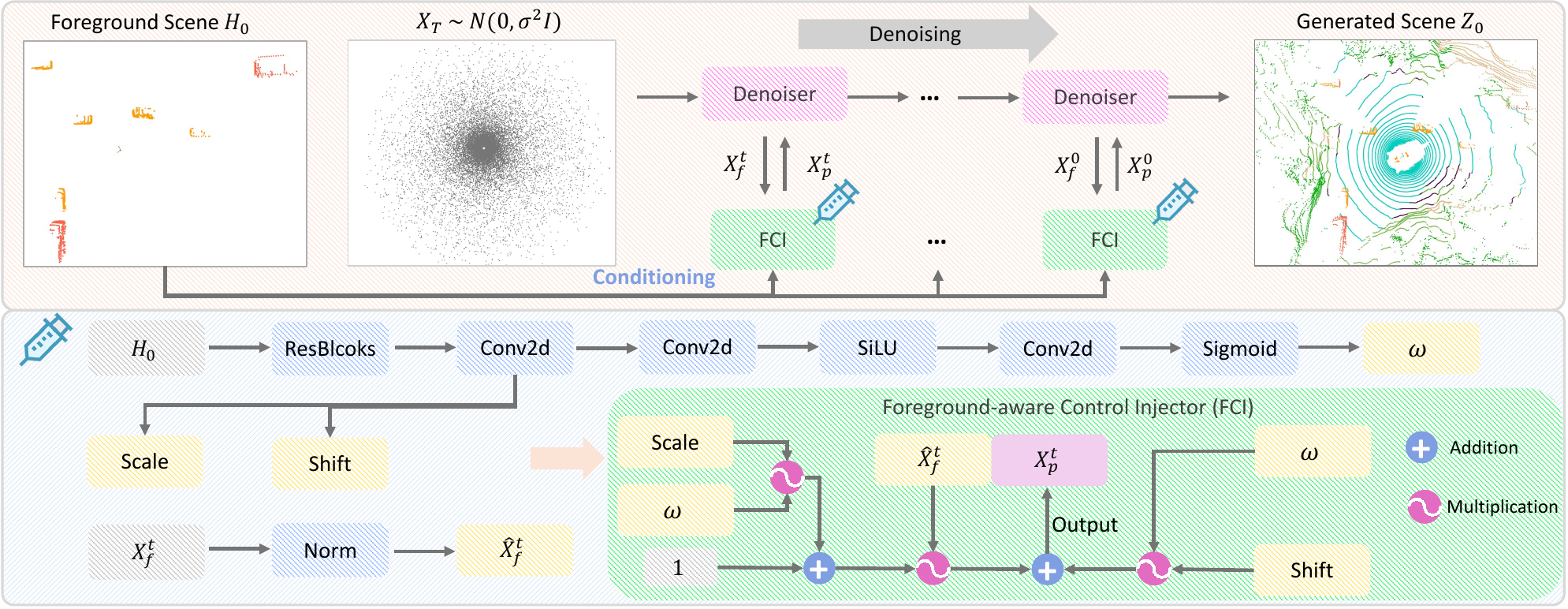}
    \vspace{-0.55cm}
    \caption{
        Architecture of our \textbf{foreground-aware LiDAR scene generation} framework. 
        \textit{Upper Part}: The diffusion-based generation process, where initial Gaussian noise $X_T \sim \mathcal{N}(0,\sigma^2I)$ is progressively denoised to generate the final scene $Z_0$, conditioned on a foreground input $H_0$ via our FCI. 
        \textit{Lower Part}: The FCI mechanism extracts features from $H_0$ and transforms them into adaptive scale and shift parameters. These modulate the intermediate features $X_f^t$ in the denoising network through channel-wise gating with attention weights $\omega$, resulting in refined features $X_p^t$ that preserve object details. This design ensures spatial coherence and semantic consistency in the generated scene $Z_0$.
    }
    \label{fig:scene_framework}
\end{figure*}

\noindent\textbf{Training Objectives.}
To promote realistic and physically consistent LiDAR layouts, we optimize a combination of geometric and diffusion-based objectives. Let $\{\hat{b}_i\}_{i=1}^M$ denote predicted layouts and $\{b_i\}_{i=1}^M$ the ground truth. We first minimize spatial collisions via a pairwise 3D IoU-based penalty:
\begin{equation}
    \mathcal{L}_{\mathrm{collision}} = \frac{1}{M} \sum_{i \neq j} \max(\mathrm{IoU}(\hat{b}_i, \hat{b}_j) - \delta, 0)~,
\end{equation}
where $\delta$ is a small tolerance threshold. To enforce alignment with ground truth geometry, we introduce an IoU-based loss:
\begin{equation}
    \mathcal{L}_{\mathrm{IoU}} = \frac{1}{M} \sum_{i} (1 - \mathrm{IoU}(\hat{b}_i, b_i))~.
\end{equation}

\noindent Additionally, we apply a standard diffusion reconstruction loss that supervises the denoising network by predicting the injected noise:
\begin{equation}
    \mathcal{L}_{\mathrm{diff}} = \mathbb{E}_{b_i, \epsilon \sim \mathcal{N}(0, I), t} \left[ \| \epsilon - \epsilon_\theta(b_i^t, t, \mathcal{C}_t) \|_2^2 \right]~.
\end{equation}

\noindent The final objective is a weighted sum of the three terms:
\begin{equation}
    \mathcal{L}_\mathrm{layout} = \lambda_1 \mathcal{L}_{\mathrm{collision}} + \lambda_2 \mathcal{L}_{\mathrm{IoU}} + \lambda_3 \mathcal{L}_{\mathrm{diff}}~,
\end{equation}
where $\lambda_1$, $\lambda_2$, and $\lambda_3$ are balancing coefficients.

\noindent\textbf{Object Synthesis.}
After generating the scene layouts, we populate them with foreground objects via either the retrieval or the generation of their LiDAR point clouds. For retrieval, an object database built from nuScenes-SG and Waymo-SG is queried to match each layout box with a semantically aligned instance of similar shape. To enhance diversity and address rare categories, we additionally train a DiT-3D-based~\cite{mo2023dit} object generator conditioned on category and layout parameters. Both pathways ensure spatial consistency and semantic coherence in object placement.

With the object synthesis strategy, our framework produces controllable LiDAR layouts by conditioning on semantic scene graphs. The integration of graph-based encoding and context-aware diffusion enables spatially coherent and semantically faithful object arrangements. These structured layouts serve as explicit priors for downstream scene synthesis, supporting precise control over object categories and spatial configurations.

\subsection{3D Scene Generation}
To enable customizable LiDAR scene generation, we generate background content conditioned on the foreground point cloud derived from the layout. This design allows users to explicitly control foreground structure while completing the scene with coherent environmental context. 
In contrast to prior diffusion-based LiDAR data generation methods~\cite{ wu2025weathergen, nakashima2024fast} that treat all points uniformly and often underrepresent sparse yet critical foreground regions, our framework ensures both semantic controllability and spatial consistency across the entire scene.

\noindent\textbf{Conditional Scene Generation.}
To condition the denoising process on structured foreground geometry, we introduce the Foreground-aware Control Injector (FCI), shown in~\Cref{fig:scene_framework}. Given the foreground point cloud $H_0$, we extract multi-scale features $\{\hat{H}_i\}_{i=1}^l$ using ResBlocks. For each injection layer, a corresponding $\hat{H}_i$ is transformed into channel-wise scale and shift parameters to modulate intermediate features. To suppress invalid regions from sparse foreground input, a binary mask $X_m \in \{0,1\}^{h \times w}$ is applied before nonlinearity at each resolution.

In addition, we introduce a gating mechanism to adaptively weight feature contributions. Specifically, each $\hat{H}_i$ is processed through a convolutional branch with dimension reduction, SiLU activation, and expansion, yielding a spatial attention map $\omega \in \mathbb{R}^{1 \times h \times w}$. The denoiser feature $X_f^t \in \mathbb{R}^{C \times h \times w}$ is then modulated as:
\begin{equation}
X_p^t = X_f^t \cdot (1 + \mathrm{scale} \cdot \omega) + \mathrm{shift} \cdot \omega~.
\end{equation}

\noindent We inject this conditioning module at multiple scales to maintain semantic awareness and spatial alignment throughout the generative hierarchy.

\noindent\textbf{Training Objectives.}
We adopt a noise reconstruction objective similar to layout generation. The primary loss minimizes the mean squared error between the predicted noise $\varepsilon_\theta(X_t, t, H_0)$ and the ground-truth noise $\epsilon$ at timestep $t$:
\begin{equation}
    \mathcal{L}_{\mathrm{scene}} = \mathbb{E}_{X, \epsilon \sim \mathcal{N}(0, I),t} \left  [\|\epsilon - \epsilon_{\theta}(X^t, t, H_0)  \|_2^2 \right]~.
\end{equation}
To emphasize fidelity in foreground regions, we introduce an auxiliary foreground loss $\mathcal{L}_{\mathrm{fore}}$, which restricts the denoising error within the masked foreground area $X_m$:
\begin{equation}
    \mathcal{L}_{\mathrm{fore}} = \mathbb{E}_{X, \epsilon \sim \mathcal{N}(0, I),t} \left  [\| (\epsilon - \epsilon_{\theta}(X^t, t, H_0)) \odot X_m  \|_2^2 \right]~.
\end{equation}
The overall training loss is a weighted sum of the two:
\begin{equation}
    \mathcal{L}_{\mathrm{cond}} = \lambda_4 \mathcal{L}_{\mathrm{scene}} + \lambda_5 \mathcal{L}_{\mathrm{fore}}~,
\end{equation}
where $\lambda_4$ and $\lambda_5$ weight each term.
\section{Experiments}
\label{sec:experiments}

\begin{table*}[t]
    \centering
    \caption{Comparisons of state-of-the-art \textbf{Layout Generation} approaches on the \textit{val} sets of \textit{nuScenes} and \textit{Waymo Open}. Metrics with $\downarrow$ indicate lower is better. All $\text{Prec}_{0.3}$, $\text{Prec}_{0.5}$ and IoU scores are given in percentage (\%).}
    \vspace{-0.2cm}
    \resizebox{0.98\linewidth}{!}{
    \begin{tabular}{r|r|cccccc|cccccc}
    \toprule
    \multirow{2}{*}{\textbf{Method}} & \multirow{2}{*}{\textbf{Venue}} & \multicolumn{6}{c|}{\textbf{nuScenes}} & \multicolumn{6}{c}{\textbf{Waymo Open}} 
    \\
    & & \textbf{RAE} $\uparrow$ & \textbf{RAD} $\uparrow$ & \textbf{CR} $\downarrow$ & $\textbf{Prec}_{0.3}$ $\uparrow$ & $\textbf{Prec}_{0.5}$ $\uparrow$ & \textbf{IoU} $\uparrow$ & \textbf{RAE} $\uparrow$ & \textbf{RAD} $\uparrow$ & \textbf{CR} $\downarrow$ & $\textbf{Prec}_{0.3}$ $\uparrow$ & $\textbf{Prec}_{0.5}$ $\uparrow$ & \textbf{IoU} $\uparrow$
    \\
    \midrule\midrule
    Graph-to-Box & ICCV'21 & $0.80$ & $0.37$ & $0.13$ & $4.69$  & $4.35$ &$17.80$& $0.74$ & $0.18$ & $0.32$  &$3.69$ &$3.39$ &$11.35$ 
    \\
    CommonScenes & NeurIPS'23 & $0.83$ & $0.45$ & $0.13$ & $4.85$ & $4.49$&  $18.71$ & $0.78$ & $0.24$ & $0.30$  & $4.44$& $4.10$ &$13.73$ 
    \\
    EchoScene & ECCV'24 & $0.91$ &$0.65$  & $0.07$ &  $5.48$ & $5.12$& $26.70$ & $\mathbf{0.91}$ & $0.55$ & $0.17$  & $5.33$ & $5.02$ & $15.67$
    \\
    \midrule
    \textbf{La La LiDAR} & \textbf{Ours} & \cellcolor{la!14}$\mathbf{0.92}$ & \cellcolor{la!14}$\mathbf{0.68}$ & \cellcolor{la!14}$\mathbf{0.06}$ & \cellcolor{la!14}$\mathbf{6.59}$ & \cellcolor{la!14}$\mathbf{6.15}$ & \cellcolor{la!14}$\mathbf{28.14}$ & \cellcolor{la!14}$\mathbf{0.91}$ & \cellcolor{la!14}$\mathbf{0.57}$ & \cellcolor{la!14}$\mathbf{0.14}$ & \cellcolor{la!14}$\mathbf{6.39}$ & \cellcolor{la!14}$\mathbf{6.08}$ & \cellcolor{la!14}$\mathbf{18.64}$
    \\
    \bottomrule
\end{tabular}}
\label{tab:comparative_layout}
\end{table*}
\begin{table}[t]
    \centering
    \caption{Comparisons of \textbf{LiDAR Scene Generation} methods on the \textit{nuScenes} dataset. Metrics with ($\downarrow$) indicate lower is better. The \textbf{MMD} metric is in $10^{-4}$.}
    \vspace{-0.2cm}
    \resizebox{\linewidth}{!}{
    \begin{tabular}{r|r|c|c|c|c}
    \toprule
    \textbf{Method} & \textbf{Venue} & \textbf{FRD}$\downarrow$ & \textbf{FPD}$\downarrow$  & \textbf{JSD}$\downarrow$ & \textbf{MMD}$\downarrow$ 
    \\
    \midrule\midrule
    LiDARGen & ECCV'22 & $549.2$ & $22.8$  & $0.04$ & $0.8$ 
    \\
    R2DM & ICRA'24 & $253.8$ & $14.4$  & $\mathbf{0.03}$ & $\mathbf{0.5}$ 
    \\
    LiDM & CVPR'24 & - & $30.8$ & $0.07$ & $3.9$
    \\
    Text2LiDAR & ECCV'24 & $953.2$  & $147.5$ & $0.09$ & $12.5$
    \\
    \midrule
    \textbf{La La LiDAR} & \textbf{Ours} & \cellcolor{la!14}$\mathbf{211.0}$ & \cellcolor{la!14}$\mathbf{9.8}$ & \cellcolor{la!14}$0.04$ & \cellcolor{la!14}$0.7$
    \\
    \bottomrule
    \end{tabular}}
\label{tab:nuscenes_gen}
\vspace{-0.1cm}
\end{table}

\subsection{Experimental Settings}
\noindent\textbf{Datasets.}
We conduct experiments on two large-scale autonomous driving datasets: nuScenes~\cite{fong2022panoptic-nuscenes, caesar2020nuscenes} and Waymo Open~\cite{sun2020waymo}. To support layout-conditioned generation, we construct two scene graph datasets, nuScenes-SG and Waymo-SG, by extracting spatial and semantic relationships among foreground objects. Construction details are provided in the Appendix.

\noindent\textbf{Evaluation Metrics.}
We evaluate LiDAR layout generation using four metrics: Relationship Accuracy Easy (RAE), Relationship Accuracy Difficult (RAD), Collision Rate (CR), and Intersection over Union (IoU). For LiDAR scene generation, we report Fréchet Range Distance (FRD), Fréchet Point Cloud Distance (FPD), Jensen-Shannon Divergence (JSD), and Maximum Mean Discrepancy (MMD). For downstream tasks, we use mean IoU (mIoU) for semantic segmentation, mean Average Precision (mAP) and nuScenes Detection Score (NDS) for 3D object detection, and Mean Absolute Error (MAE) for scene completion. Definitions of all metrics are provided in the Appendix.

\subsection{Comparative Study}
\noindent\textbf{LiDAR Layout Generation.}
\Cref{tab:comparative_layout} shows that our method outperforms previous methods across all metrics on both datasets. On nuScenes, we achieve higher precision ($6.59\%$ and $6.15\%$) and IoU ($28.14\%$), with better relationship accuracy (RAE: $0.92$, RAD: $0.68$) and lower collision rate ($0.06$). Similar improvements are observed on Waymo Open. These results demonstrate the effectiveness of our approach in generating spatially coherent foreground layouts.

\begin{figure*}[t]
    \centering
    \includegraphics[width=0.99\textwidth]{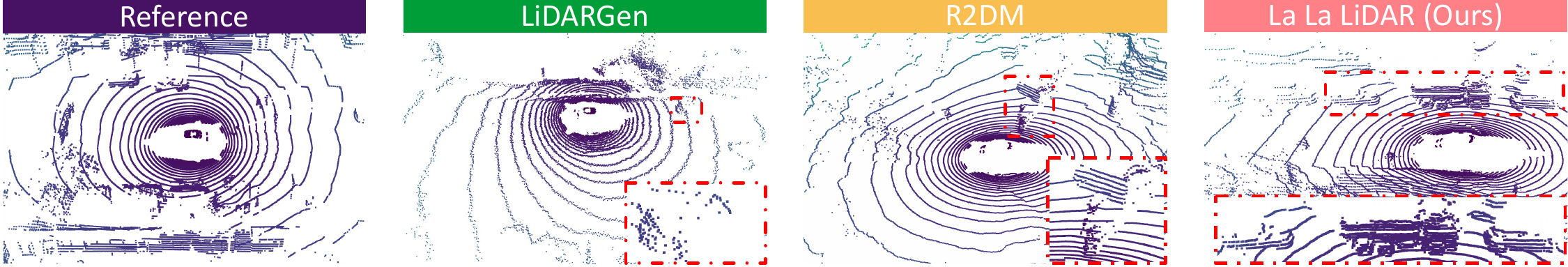}
    \vspace{-0.1cm}
    \caption{Qualitative comparisons of La La LiDAR against state-of-the-art LiDAR scene generation approaches on the nuScenes dataset. From left to right: Reference (ground truth), LiDARGen, R2DM, and our method.}
    \label{fig:vis_compare}
\end{figure*}

\noindent\textbf{LiDAR Scene Generation.}
Our method achieves state-of-the-art performance on LiDAR scene generation, as shown in \Cref{tab:nuscenes_gen}. La La LiDAR fulfills the lowest FPD ($9.8$) and FRD ($211.0$), outperforming LiDARGen ($22.8$, $549.2$) and R2DM ($14.4$, $253.8$). These results indicate that our model generates LiDAR samples with both high geometric fidelity and semantic alignment. \Cref{fig:vis_compare} presents qualitative comparisons against LiDARGen and R2DM. LiDARGen introduces artifacts in the background and fails to maintain structural consistency. R2DM exhibits a smoother global layout but lacks foreground sharpness and coherent fusion with surrounding regions. In contrast, our model produces scenes with detailed foreground geometry and seamless background integration. This visual fidelity benefits from our FCI mechanism, which ensures accurate conditioning and structure-aware denoising. Additional visual evidence is provided in the Appendix. These qualitative and quantitative results together validate the effectiveness of our method in generating high-quality, controllable LiDAR scenes.

\begin{figure*}[t]
    \centering
    \includegraphics[width=\textwidth]{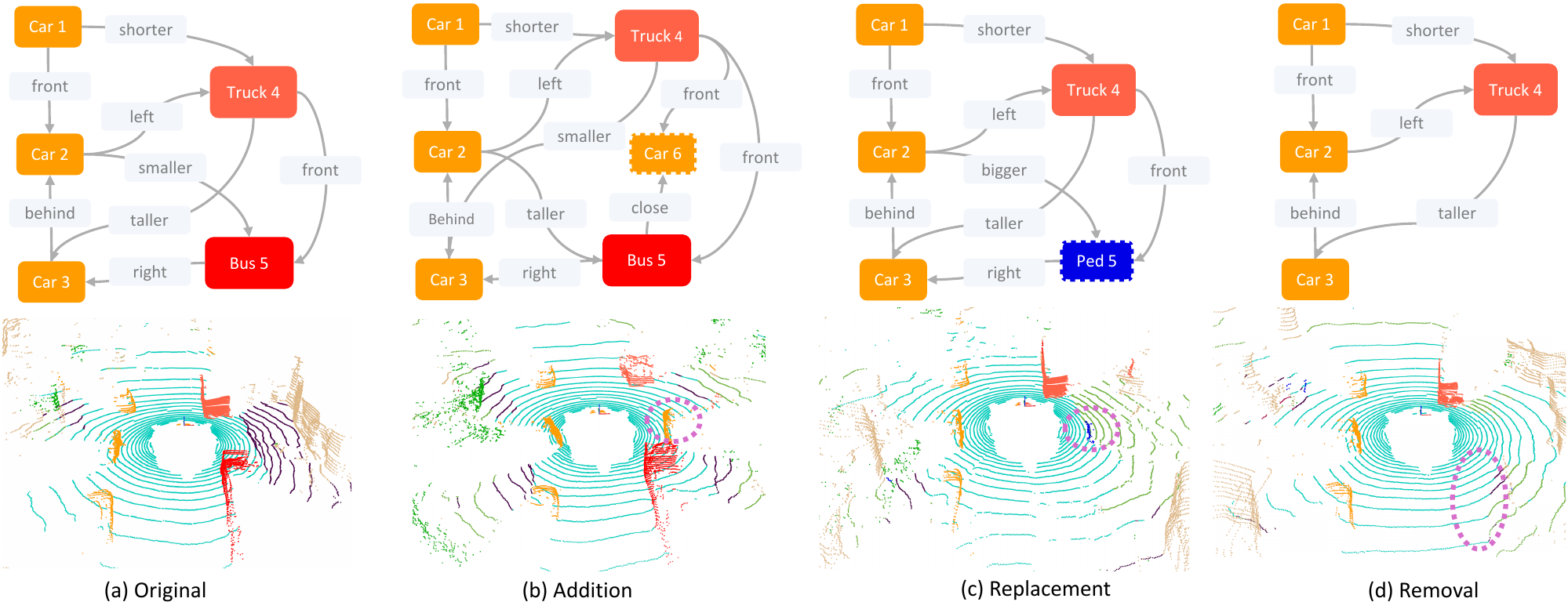}
    \vspace{-0.5cm}
    \caption{Controllable scene synthesis via graph-level editing. \textbf{Top:} input scene graphs with object nodes and relational edges. \textbf{Bottom:} generated LiDAR scenes. Edge visualizations are partially omitted for clarity. LiDAR point clouds are colorized via a pre-trained LiDAR segmentation network~\cite{tang2020searching}.}
    \label{fig:vis_manipulation}
\end{figure*}

\noindent\textbf{LiDAR Semantic Segmentation.}
Moreover, our method could be tamed as a generative data augmentation strategy for LiDAR segmentation, where generated samples (pseudo-labeled using a pre-trained SPVCNN) are combined with limited real data during training. As shown in \Cref{tab:nusc_seg}, our method consistently improves performance under sparse-label settings, achieving 65.1\% mIoU with MinkUNet and 65.4\% with SPVCNN at 1\% supervision, outperforming both R2DM and the semi-supervised method LaserMix~\cite{kong2023laserMix}. These results demonstrate the effectiveness of our generated scenes in enhancing label efficiency for LiDAR segmentation.

\noindent\textbf{3D Object Detection.} 
We further investigate the utility of La La LiDAR in the detection task by augmenting the training data with generated samples. As shown in \Cref{tab:detection}, our method consistently improves performance across various data regimes, even with significant gains in low-data settings. These findings validate the effectiveness of our approach in enhancing detector training when real annotations are scarce.

\noindent\textbf{Scene Completion.}
Furthermore, we evaluate our framework on LiDAR completion using a Repaint~\cite{lugmayr2022repaint} where every fourth beam is preserved. As shown in \Cref{tab:completion}, we outperform both interpolation and diffusion-based approaches across all metrics, demonstrating superior structural and semantic coherence during completion.

\begin{table}[t]
    \centering
    \caption{
        Downstream application of \textit{\textbf{La La LiDAR}} for the \textbf{3D Semantic Segmentation} task on the \textit{val} set of \textit{nuScenes}. 
    }
    \vspace{-0.2cm}
    \resizebox{\linewidth}{!}{
    \begin{tabular}{c|r|r|cccc}
    \toprule
    \multirow{2}{*}{\textbf{\#}} & \multirow{2}{*}{\textbf{Method}} & \multirow{2}{*}{\textbf{Venue}} & \multicolumn{4}{c}{\textbf{nuScenes}} 
    \\
    & & & \textbf{1\%} & \textbf{10\%} & \textbf{20\%} & \textbf{50\%} 
    \\
    \midrule\midrule
    \multirow{5}{*}{\rotatebox{90}{\textbf{MinkU.}}}
    & \textit{Sup.-only} & - & $58.3$ & $71.0$ & $73.0$ & $75.1$
    \\
    & MeanTeacher & NeurIPS'17 & $60.1$ & $71.7$ & $73.4$ & $75.2$
    \\
    & LaserMix & CVPR'23 & $62.8$ & $73.6$ & $74.8$ & $76.1$
    \\
    & R2DM & ICRA'24 & $64.1$ & $73.0$ & $74.3$ & $75.9$
    \\
    \cmidrule{2-7}
    & \textbf{La La LiDAR} & \textbf{Ours} & \cellcolor{la!14}$\mathbf{65.1}$ & \cellcolor{la!14}$\mathbf{73.8}$ & \cellcolor{la!14}$\mathbf{75.4}$ & \cellcolor{la!14}$\mathbf{76.2}$
    \\
    \midrule
    \multirow{5}{*}{\rotatebox{90}{\textbf{SPVCNN}}}
    & \textit{Sup.-only} & - & $57.9$ & $71.7$ & $73.0$ & $74.6$
    \\
    & MeanTeacher & NeurIPS'17 & $59.4$ & $72.5$ & $73.1$ & $74.7$
    \\
    & LaserMix & CVPR'23 & $63.2$ & $\mathbf{74.1}$ & $74.6$ & $75.8$
    \\
    & R2DM & ICRA'24 & $64.6$ & $72.7$ & $74.2$ & $75.4$
    \\
    \cmidrule{2-7}
    & \textbf{La La LiDAR} & \textbf{Ours} & \cellcolor{la!14}$\mathbf{65.4}$ & \cellcolor{la!14}$74.0$ & \cellcolor{la!14}$\mathbf{75.0}$ & \cellcolor{la!14}$\mathbf{76.3}$
    \\\bottomrule
    \end{tabular}}
    \label{tab:nusc_seg}
\end{table}
\begin{table}[t]
    \centering
    \caption{
        Downstream application of \textit{\textbf{La La LiDAR}} for the \textbf{3D Object Detection} task on the \textit{val} set of \textit{nuScenes}. 
    }
    \vspace{-0.2cm}
    \resizebox{\linewidth}{!}{
    \begin{tabular}{c|r|cc|cc|cc|cc}
    \toprule
    \multirow{2}{*}{\textbf{\#}} & \multirow{2}{*}{\textbf{Method}} & \multicolumn{2}{c|}{\textbf{1\%}} & \multicolumn{2}{c|}{\textbf{5\%}} & \multicolumn{2}{c|}{\textbf{10\%}} & \multicolumn{2}{c}{\textbf{20\%}} 
    \\
    & & \textbf{mAP} & \textbf{NDS} & \textbf{mAP} & \textbf{NDS} & \textbf{mAP} & \textbf{NDS} & \textbf{mAP} & \textbf{NDS} 
    \\
    \midrule\midrule
    \multirow{4.5}{*}{\rotatebox{90}{\textbf{Center.}}}
    & \textit{Sup.-only} & $24.7$ & $31.2$ & $36.7$ & $40.1$ & $40.9$ & $43.0$ & $40.7$ & $42.9$ 
    \\
    & R2DM & $26.4$ & $32.1$ & $36.9$ & $41.2$ & $41.0$ & $43.8$ & $41.9$ & $45.0$ 
    \\
    & Text2LiDAR & $26.0$ & $31.8$ & $36.8$ & $40.7$ & $41.2$ & $43.9$ & $41.4$ & $44.9$ 
    \\
    \cmidrule{2-10}
    & \textbf{La La LiDAR} & \cellcolor{la!14}$\mathbf{27.0}$ & \cellcolor{la!14}$\mathbf{32.6}$ & \cellcolor{la!14}$\mathbf{37.0}$ & \cellcolor{la!14}$\mathbf{41.6}$ & \cellcolor{la!14}$\mathbf{41.4}$ & \cellcolor{la!14}$\mathbf{44.4}$ & \cellcolor{la!14}$\mathbf{42.5}$ & \cellcolor{la!14}$\mathbf{45.6}$ 
    \\
    \midrule
    \multirow{4.5}{*}{\rotatebox{90}{\textbf{SEC.}}}
    & \textit{Sup.-only} & $0.8$ & $19.4$ & $29.2$ & $36.5$ & $34.5$ & $41.2$ & $34.3$ & $41.3$ 
    \\
    & R2DM & $13.1$ & $21.2$ & $32.7$ & $39.4$ & $36.2$ & $42.1$ & $39.7$ & $44.5$ 
    \\
    & Text2LiDAR & $15.0$ & $25.4$ & $31.3$ & $38.5$ & $34.6$ & $41.4$ & $37.1$ & $43.3$ 
    \\
    \cmidrule{2-10}
    & \textbf{La La LiDAR} & \cellcolor{la!14}$\mathbf{15.4}$ & \cellcolor{la!14}$\mathbf{25.8}$ & \cellcolor{la!14}$\mathbf{33.1}$ & \cellcolor{la!14}$\mathbf{40.0}$ & \cellcolor{la!14}$\mathbf{36.6}$ & \cellcolor{la!14}$\mathbf{42.6}$ & \cellcolor{la!14}$\mathbf{39.9}$ & \cellcolor{la!14}$\mathbf{44.9}$ 
    \\
    \bottomrule
    \end{tabular}
    }
    \label{tab:detection}
    \vspace{-0.1cm}
\end{table}
\begin{table}[t]
    \centering
    \caption{Downstream application of \textit{\textbf{La La LiDAR}} for the \textbf{LiDAR Completion} task on the \textit{nuScenes}. \textbf{Inter.}: Interpolation methods. \textbf{Diff.}: Diffusion-based methods.}
    \vspace{-0.2cm}
    \resizebox{\linewidth}{!}{
    \begin{tabular}{c|r|c|c|c}
    \toprule
    \multirow{2}{*}{\textbf{\#}} & \multirow{2}{*}{\textbf{Method}} & \textbf{MAE} $\downarrow$ & \textbf{MAE} $\downarrow$ & \textbf{IoU} $\uparrow$ 
    \\
    & & {\small (Range)} & {\small (Reflectance)} & {\small (Semantics)}
    \\
    \midrule\midrule
    \multirow{3}{*}{\rotatebox{90}{\textbf{Inter.}}}
    & Nearest Neighbor &$4.00$ & $8.16$& $25.11$
    \\
    & Bi-Linear &$4.27$ & $8.59$ & $31.27$
    \\
    & Bi-Cubic & $4.59$ &  $9.10$& $25.23$
    \\
    \midrule
    \multirow{2}{*}{\rotatebox{90}{\textbf{Diff.}}} 
    & R2DM & $2.64$ & $5.63$ & $64.44$
    \\
    & \textbf{La La LiDAR} & \cellcolor{la!14}$\mathbf{2.45}$ & \cellcolor{la!14}$\mathbf{5.10}$  & \cellcolor{la!14}$\mathbf{66.90}$
    \\
    \bottomrule
    \end{tabular}}
\label{tab:completion}
\vspace{-0.1cm}
\end{table}

\subsection{Ablation Study}
\noindent\textbf{Analysis of Key Design Choices for Layout Generation.}
We conduct ablation studies in \Cref{tab:layout_ablation} to assess three key designs of our layout generation framework: semantic graph encoding, conditioning mechanism, and geometric consistency losses. First, excluding semantic features (row a \textit{vs.} d) markedly reduces relationship accuracy (RAD: $0.68$ $\to$ $0.59$) and spatial precision (IoU: $28.14\%$ $\to$ $23.64\%$), highlighting the contribution of scene-level semantics. Second, replacing cross-attention with naive concatenation (row b \textit{vs.} d) leads to degraded RAD ($0.68$ $\to$ $0.63$) and object alignment (Prec$_{0.5}$: $6.15\%$ $\to$ $4.82\%$), indicating that attention mechanisms better capture relational structure. Finally, ablating our geometric loss terms (row c \textit{vs.} d) increases the collision rate and further reduces IoU, confirming their effectiveness in promoting physically consistent layouts.

\noindent\textbf{Ablation on Foreground Mask and Conditioning Strategy.} 
We ablate two key components of our scene generation framework: the foreground validity mask $X_m$ and the FCI mechanism. Removing $X_m$ (row c in \Cref{tab:lidar_ablation}) leads to a notable degradation across all metrics (\textit{e.g.}, FPD: $9.8 \to 19.5$), confirming the necessity of suppressing invalid sparse regions during feature modulation. To evaluate conditioning strategies, we compare addition (row d), concatenation (row e), and cross-attention alone (row f). While both addition and concatenation outperform the baseline, cross-attention performs poorly (FPD: $27.5$), likely due to unstable modulation. Our final model (row g), which integrates $X_m$ and the FCI module, achieves the best overall performance, demonstrating that structured and spatially aligned conditioning is essential for controllable and coherent scene synthesis.

\begin{table}[t]
    \centering
    \caption{Ablation study of the outdoor LiDAR layout generation on the \textit{nuScenes} dataset. All metrics ($\downarrow$) indicate lower is better. \textit{SG} is the semantic graph, and \textit{OI} refers to the combination of $\mathcal{L}_\mathrm{collosion}$ and $\mathcal{L}_\mathrm{IoU}$.}
    \vspace{-0.2cm}
    \resizebox{\linewidth}{!}{
    \begin{tabular}{c|l|c|c|c|c|c|c}
    \toprule
    \textbf{\#} & \textbf{Configuration} & \textbf{RAE}$\uparrow$ &  \textbf{RAD}$\uparrow$& \textbf{CR} $\downarrow$ & $\textbf{Prec}_{0.3}$$\uparrow$ & $\textbf{Prec}_{0.5}$$\uparrow$ & \textbf{IoU} $\uparrow$
    \\
    \midrule\midrule
    a & Ours \textit{w/o} SG & $0.91$ & $0.59$ &  $0.08$& $5.01$  & $4.76$  & $23.64$
    \\

    b & Ours \textit{w/} Concat & $\mathbf{0.92}$ & $0.63$  & $0.07$ & $5.15$  & $4.82$ &$25.89$
    \\
    c & Ours \textit{w/o} OI  & $\mathbf{0.92}$  & $0.62$ & $0.08$& $5.96$ & $5.50$ & $26.84$
    \\
    \midrule
    \textbf{d} & \textbf{The Full Framework} & \cellcolor{la!14}$\mathbf{0.92}$ & \cellcolor{la!14}$\mathbf{0.68}$ & \cellcolor{la!14}$\mathbf{0.06}$ & \cellcolor{la!14}$\mathbf{6.59}$ & \cellcolor{la!14}$\mathbf{6.15}$ & \cellcolor{la!14}$\mathbf{28.14}$
    \\\bottomrule
    \end{tabular}}
    \label{tab:layout_ablation}
\end{table}
\begin{table}[!t]
    \centering
    \caption{Ablation study of our LiDAR scene generation framework on the \textit{nuScenes} dataset. The MMD is in $(10^{-4})$. All metrics ($\downarrow$) indicate that lower is better. FM refers to foreground mask and CA denotes cross-attention.}
    \vspace{-0.2cm}
    \resizebox{\linewidth}{!}{
    \begin{tabular}{c|l|c|c|c|c}
    \toprule
    \textbf{\#} & \textbf{Configuration} & \textbf{FRD}$\downarrow$ & \textbf{FPD}$\downarrow$  & \textbf{JSD}$\downarrow$ & \textbf{MMD}$\downarrow$
    \\
    \midrule\midrule
    a & Baseline & $253.79$  & $14.41$  & $ \mathbf{0.034}$  &  $\mathbf{0.54}$ 
    \\
    b & Box layout condition & $244.60$  & $14.22$  & $ 0.051$  &  $1.24$ 
    \\
    c & Ours \textit{w/o} {FM} & $301.34$  & $19.45$ & $0.048$  &  $1.66$
    \\
    d & Ours \textit{w/} {Add} condition & $235.37$ & $\mathbf{9.38}$ & $0.047$ &  $1.38$
    \\
    e & Ours \textit{w/} {Concat} condition & $232.15$ & $9.65$ & $0.047$ & $1.30$ 
    \\
    f & Ours \textit{w/} {CA} condition & $461.85$ & $27.47$ & $0.091$ &  $3.77$ 
    \\
    \midrule
    \textbf{g} & \textbf{The Full Framework} & \cellcolor{la!14}$\mathbf{211.03}$ & \cellcolor{la!14}$9.82$ & \cellcolor{la!14}$0.039$ & \cellcolor{la!14}$0.74$
    \\\bottomrule
    \end{tabular}}
\label{tab:lidar_ablation}
\vspace{-0.1cm}
\end{table}

\noindent\textbf{Advantages of Scene Graph Conditioning over Box Layout.} 
To evaluate the effectiveness of structured relational priors, we compare our scene graph-based conditioning with a baseline that uses only 3D box layouts (row b in \Cref{tab:lidar_ablation}). Despite providing geometric cues, the box-only representation underperforms our method in both semantic consistency (FRD: $244.6$ $\to$ $211.0$) and geometric fidelity (FPD: $14.2$ $\to$ $9.8$), underscoring the importance of explicit relational structure in guiding scene generation. Beyond quantitative performance, scene graphs offer several key advantages:  
(1) They encode inter-object relationships (\textit{e.g.}, \textit{car behind pedestrian}) that box layout cannot represent, enabling richer spatial reasoning;  
(2) They support hierarchical control, allowing users to specify both object presence and their relational configuration;  
(3) They serve as structured, extensible priors that integrate naturally with downstream autonomy modules such as planning or forecasting. These insights highlight the advantages of structured relational priors for controllable LiDAR scene generation.

\noindent\textbf{Graph-Driven Customization of LiDAR Scene Generation.} \Cref{fig:vis_manipulation} illustrates the controllable generation capability of our framework through scene graph editing. Starting from an initial graph (a), we apply three types of modifications: (b) insert a new object with relational edges, (c) substitute an object category, and (d) remove an object node. Each operation induces coherent changes in the generated LiDAR scenes, demonstrating the model’s ability to respond to structured semantic edits. These results demonstrate the flexibility of our approach in enabling fine-grained, relation-aware control over scene composition and structure.
\section{Conclusion}
\label{sec:conclusion}

In this work, we propose \textbf{La La LiDAR}, a novel layout-guided generative framework for controllable 3D LiDAR scene generation. Our approach enables controllable control over object placement and spatial relationships, addressing the limitations of existing unconditional LiDAR generation methods. To support structured LiDAR generation, we introduce Waymo-SG and nuScenes-SG, two large-scale LiDAR scene graph datasets, along with new evaluation metrics specifically designed for LiDAR layout generation. Extensive experiments demonstrate that our framework achieves state-of-the-art performance in both LiDAR generation and downstream perception tasks, validating its effectiveness for real-world applications.

\bibliography{main}

\end{document}